\documentclass[runningheads]{llncs}
\usepackage{amsmath}
\usepackage{amsfonts}
\usepackage{amssymb}
\usepackage{graphicx}
\usepackage[utf8]{inputenc}
\usepackage[ruled]{algorithm2e}

\begin{document}

\title{Fine-tuning ClimateBert transformer with ClimaText for the disclosure analysis of climate-related financial risks}
\titlerunning{Fine-tuning ClimateBert transformer for the analysis of climate-related risks}
\author{Eduardo C. Garrido-Merchán, Cristina González-Barthe, María Coronado Vaca}
\authorrunning{Eduardo C. Garrido-Merchán et al.}
\date{March 2023}

\institute{Universidad Pontificia Comillas, Madrid, Spain \\
\email{ecgarrido@icade.comillas.edu, 201702658@alu.comillas.edu, mcoronado@icade.comillas.edu}}

\maketitle

\begin{abstract}
In recent years there has been a growing demand from financial agents, especially from particular and institutional investors, for companies to report on climate-related financial risks. A vast amount of information, in text format, can be expected to be disclosed in the short term by firms in order to identify these types of risks in their financial and non financial reports, particularly in response to the growing regulation that is being passed on the matter. To this end, this paper applies state-of-the-art NLP techniques to achieve the detection of climate change in text corpora. We use transfer learning to fine-tune two transformer models, BERT and ClimateBert -a recently published DistillRoBERTa-based model that has been specifically tailored for climate text classification-. These two  algorithms are based on the transformer architecture which enables learning the contextual relationships between words in a text. We carry out the fine-tuning process of both models on the novel “ClimaText” database, consisting of data collected from Wikipedia, 10K Files Reports and web-based claims.  Our text classification model obtained from the ClimateBert fine-tuning process on ClimaText, outperforms the models created with BERT and the current state-of-the-art transformer in this particular problem. Our study is the first one to implement on the ClimaText database the recently published ClimateBert algorithm. Based on our results, it can be said that ClimateBert fine-tuned on ClimaText is an outstanding tool within the NLP pre-trained transformer models that may and should be used by investors, institutional agents and companies themselves to monitor the disclosure of climate risk in financial reports, as well as in other textual sources, such as policies, new legislation or directives passed on this matter.  In addition, our transfer learning methodology is cheap in computational terms, thus allowing any organization to perform it.  
\end{abstract}

\keywords{environmental disclosure, climate-related financial risks, climate change, ClimaText, ClimateBert, Natural Language Processing (NLP), text classification, green finance, sustainability.}

\section{Introduction}
In recent years there has been a growing demand from financial agents, especially from particular and institutional investors, for companies to report on climate-related financial risks. There is also a recent trend towards mandatory rather than voluntary reporting worlwide. Thus, a vast amount of information, in text format, can be expected to be disclosed in the short term by firms in order to identify these types of risks in their financial and non financial reports, particularly in response to the growing regulation that is being passed on the issue.   

The Financial Stability Board (FSB) created the Task-Force on Cimate-Related Financial Disclosures (TCFD) in 2015 at the request of the G-20 leaders, to improve and increase disclosure of climate-related financial information. In 2017, the TCFD released climate-related financial disclosures recommendations designed to help companies provide better information to support informed capital allocation \cite{board2017recommendations}. The latest TCFD status report describes the steady increase in climate-related financial disclosures since 2017. As of November 2022, the number of TCFD supporters surpasses 4,000 organizations from more than 100 countries with a combined market capitalization of \$27 trillion \cite{board2022task}. Many international organizations have developed various frameworks to help companies in their discloure of environmental, social and governance (ESG) issues, including the GRI (Global Reporting Initiative), CDP (Carbon Disclosure Project), SASB (Sustainability Accounting Standards Board), and some others. 

The European Union (EU) became the first jurisdiction to require its largest companies to provide information on environmental, personnel and human rights, and sustainability issues \cite{steffen2021comparative}. The European Non-Financial Reporting Directive (European NFRD, 2014) obliges companies subject to it to prepare a Report of Non-Financial Information (NFIR), through which to provide this information Many EU countries have already transpose this NFR Directive into their national law (Austria, Germany, Hungary, Ireland, Italy, The Netherlands, Poland, Slovenia, Spain, etc.). The EU’s Sustainable Finance Disclosure Regulation (SFDR) came into force on March 10th, 2021 (European SFDR, 2019) and EU Taxonomy disclosures have been required starting on Dec 31st, 2021 (European Taxonomy Regulation, 2020). On November 16, 2022, the European Financial Reporting Advisory Group (EFRAG) approved the updated versions of the European Sustainability Reporting Standards (ESRS). The new standards are part of Europe’s Corporate Sustainability Reporting Directive (CSRD) which was approved on November 28. 2022. The CSRD is expected to enter into force for reporting year 2024, with first submissions due in 2025 (European CSRD, 2022). It aims to strengthen sustainability reporting requirements under the existing NFRD to improve corporate accountability, as well as the quality, consistency, and comparability of information disclosed. This new Directive modernises and strengthens the rules about the social and environmental information that companies have to report. A broader set of large companies, as well as listed SMEs, will now be required to report on sustainability – approximately 50 000 companies in total. In France, insitutional investors already face their own disclosure requirements on climate risk management, and how they are contributing to the transition energetic. Article 173 of the French Energy Transition Law , came into force in January 2016. It has been pioneer to introduce the obligation for institutional investors to inform on the carbon footprint of their investments. Investors with a balance of more than 500 million euros were obliged to submit their first reports for 2016, in June 2017 (French Law on Energy Transition and Green Growth, 2015). The United Kingdom (UK) has become first G20 country to make it mandatory for Britain’s largest businesses to disclose their climate-related risks and opportunities, in line with TCFD recommendations, with new legislation that came into force from April 2022 (GOV UK, 2022). New Zealand also makes climate-related disclosures mandatory, from January 2023, for large publicly listed companies, insurers, banks, non-bank deposit takers and investment managers (New Zealand Ministry for the Environment, 2021). The United States (US) Securities and Exchange Commission (SEC) first addressed disclosure of material environmental issues in the early 1970s (SEC, 2010).  The SEC has released a proposal in March 2022 that would require companies to disclose climate-related disclosures for investors (SEC, 2022).  

On 3 November 2021, at COP26, the International Financial Reporting Standards (IFRS) Foundation announced the creation of the International Sustainability Standards Board (ISSB) to deliver a global baseline of sustainability-related disclosures to meet capital market needs of high-quality, transparent, reliable and comparable reporting by companies on climate and other environmental, social and governance (ESG) issues. In March 2022, the ISSB published Exposure Draft IFRS S2 Climate-related Disclosures and, by july 2022, received more than 600 responses to it (IFRS, 2022). The ISSB is finalising requirements for an entity to disclose information about its climate-related risks and opportunities and expects to issue an IFRS Sustainability Disclosure Standard around the end of Q2 2023.  The ISSB’s work is backed by the G7, the G20, IOSCO (International Association of Securities Commissions), the FSB, African Finance Ministers and by Finance Ministers and Central Bank Governors from over 40 jurisdictions.  

According to the World’s Economic Forum Global Risk Report 2023, climate and environmental risks are the core focus of global risks perceptions over the next decade – and are the risks for which we are seen to be the least prepared. All the six environmental risks feature in the top 10 risks over the next 10 years, with “failure to mitigate climate change”, “failure of climate-change adaptation”, “natural disasters and extreme weather events”, and “biodiversity loss and ecosystem collapse” having ranked as top four most critical risks (World Economic Forum, 2023, p.6).  

In this context, Carney (2015) already stated in his speech “Breaking the tragedy of the horizon - climate change and financial stability” that climate-related financial risks were being belittled by financial agents, bearing these the risk of triggering the next financial crisis. 

Climate change as a financial risk is a growing concern among several financial agents seeking ways of assessing and measuring this threat. To this effect, Ilhan et al. \cite{ilhan2021climate} carried out a survey among institutional investors of which one third of the respondents hold an executive position at their institution and 11\% work for firms that amount more than \$100bn in assets . This survey showed came that 51\% of these investors believe that reporting on climate risk is as important as financial reporting and almost one third considered even more relevant the first. There is a growing demand from investors for climate risk reporting as well as a higher disposition to engage with organizations that do carry out such a disclosure. Our study is framed within this context of both, growing awareness and demand for financial reporting of climate-related risks, and of recent trend towards mandatory rather than voluntary reporting worldwide. 

The growing importance of these disclosures, with their intrinsic characteristic of heterogeneity and dispersed features, make the task of studying and analysing these type of financial and non financial reports worthy of automation. As a result, in recent years, a growing literature has emerged that relies on AI for the identification of climate-related information \cite{sautner2020firm,meddeb2022counteracting,kolbel2020ask,bingler2022cheap,sietsma2022global,webersinke2021climatebert,friederich2021automated,hsu2021diverse,luccioni2019using,luccioni2020analyzing,biesbroek2020machine,alashri2016climate}, among others.  

Traditional NLP (Natural Language Processing) techniques such as the “Bag-of-Words" (BoW) approaches have been predominant \cite{donner2016measuring,loughran2016textual,cody2015climate,sham2022climate}, although this technique has the significant  drawback of assuming that the words in the text are independent of one another what is evidently false. Further literature has relied on word embeddings technology \cite{luccioni2019using} and, even though they are an improvement on the use of BoW, the terminology regarding climate change is likely to change drastically in relation to the sources which are used in the research, so this NLP technique does not offer a completely satisfactory answer either. Furthermore, it should not be forgotten that the vector representation does not take into account the context of the sentence and therefore negations are not considered by this approach. As a result, both methods are very far from understanding sentences in their full extent \cite{kolbel2022ask}.  

However, the field is experiencing an enormous revolution since the implementation of transformer models has been possible due to the rise of computing \cite{meddeb2022counteracting} – using a CamemBERT Transformer-based model, a transformer french language model, they  classify french news on climate change, identifying those which are fake ones- \cite{kolbel2022ask,bingler2022cheap,webersinke2021climatebert,friederich2021automated,luccioni2020analyzing}.  Transformers are large language models that capture the dependencies between words, that are encoded in word embeddings whose space represents the meaning of the words. Specifically, transformers’ empirical results dramatically outperform the classical pipeline of machine learning models with a bag-of-words representation of the most common and relevant words of the texts according to algorithms such as TF-IDF (Term Frequency-Inverse Document Frequency) \cite{gonzalez2020comparing}. But transformers can only be accurately estimated by supercomputing centers or organizations that have lots of computing power, being impossible to be trained from zero by small organizations or businesses for small supervised learning tasks. To avoid this limitation for its application, transfer learning can be used to fine-tune a transformer trained in a similar task to the one that needs to be solved. The fine-tuning process adapts the behaviour of the transformer to the particular task to be solved and it is a cheap in computational terms, thus allowing any organization to perform it. Moreover, the results that a fine-tuned transformer can deliver outperform classical methodologies, or small models trained from zero \cite{gonzalez2020comparing,kolbel2022ask,friederich2021automated}. For this reason, we propose fine-tuning the recently published pretained transformer ClimateBert that has been specifically tailored for climate text classification \cite{webersinke2021climatebert} on the database ClimaText \cite{varini2020climatext}. The purpose of this paper is to allow better analysis and identification of the increasing amount of climate-related risk disclosure by companies; thus, to further contribute to the literature on the detection of the topic of climate change in text corpora, particularly in order to encourage the identification by economic agents of climate-related financial risks disclosure of firms. 

This article contributes to the literature on environmental disclosure in several ways: 

Our research entails, to the best of our kowledge, the first ever use of the recently published ClimateBert algorithm on the ClimaText database, achieving, as developed throughout the paper, better results than those obtained so far from the implementation of other algorithms on ClimaText by Varini et al \cite{varini2020climatext}. Also, our study marks the first application of the ClimaText database, having been previously used only by its own authors \cite{varini2020climatext} (one more non published paper, includes this reference in its bibliography but does not use it in its research, instead applying another database). 

Since its publication in October 2021, the pretained transformer model ClimateBert has only been applied so far in 6 other papers which have not been published yet (two of which are from ClimateBert's original authors): \cite{hershcovich2022towards,bingler2022cheapb,sietsma2022global,brie2022mandatory,fard2022climedbert,yu4308287climatebug}. Hershcovich et al. \cite{hershcovich2022towards} analyze ClimateBert only regarding its energy consumption, in a context of awareness about the environmental impact that NLP pre-trained models present. Focusing on the policies being adopted by governments around the world, Sietsma et al. \cite{sietsma2022global} identify ClimateBert as a tool that allows real-time tracking of adaptation progress. Yu et al. \cite{yu4308287climatebug} present a database, climateBUG, that provides a framework for detecting implicit information about how banks disclose their climate change-related activities. They use ClimateBERT to check the classification accuracy of this dataset, the study results suggesting that ClimateBert outperforms FinBert and Bert uncased due to its higher domain accuracy, although the best performance was obtained by climateBUG-LM (Language Model). Brié et al. \cite{brie2022mandatory} apply ML techniques to quantify the content of climate disclosure in relation to the European Non-Financial Reporting Directive (NFRD), studying the change in attitude of companies in their climate-risk reporting. To do this, they use ClimateBERT to accurately identify whether companies are disclosing information related to their risk as a result of climate change. Fard et al. \cite{fard2022climedbert} seek to create a language model in such a way that it allows solving different tasks, such as detecting similarities between concepts related to climate and health, checking facts, extracting relationships of health effects in the generation of political texts, etc. Thus, they create CliMedBERT, which is based on ClimateBERT and diseaseBERT. The surprising conclusion reached by Bingler et al. \cite{bingler2022cheapb} in their application of ClimateBert to TFCD reports is particularly striking. In the face of the expected growth of disclosure of climate-related risks by TFCD supporting companies, the study concludes that said support is mostly cheap talk as firms cherry pick in order to report primarily non-material climate risk information. They continue this study with a new paper, in which Bingler et al. \cite{bingler2022cheapb} conclude that “institutional ownership, targeted institutional investor engagement, materiality and downside risk disclosures are associated with less cheap talk”.   

Numerous stakeholders rely on climate-related corporate disclosures to regularly evaluate risks, impacts, consequences and opportunities linked to their investments, lending or insurance portfolios. In this regards, the present study will be relevant and useful to investors and financial agents as it will help establish whether firms are reporting and disclosing climate-related information considered relevant within their industry sector. Likewise, governmental authorities, regulators or policy makers might also be interested in the present project to assess a specific industry’s or firm’s current state in relation to climate risk disclosure.  

Various studies have analyzed various aspects of environmental, social and governance (ESG) disclosure. We refer to  Ellili et al. \cite{ellili2022bibliometric} for a bibliometric and systematic review of this literature. They identify four most relevant areas in ESG disclosure research up to date: corporate social responsibility (CSR), corporate strategy, financial performance and environmental economics. Our study and obtained results, to the extent that we make possible a better identification and analysis of the climate-related information disclosure, it will also allow advancing in the research on whether the quantity and quality of a company’s ESG disclosure affects a great variety of this firm’s variables included in these four big areas identified by Ellili et al. \cite{ellili2022bibliometric}. Thus, we extend our paper’s target audience to all these strands of the literature related to the ESG disclosure.  

The remainder of this paper is organized as follows. In section 2, we first show the fundamentals of transfer learning to understand the motivation behind choosing this methodology. Then we describe the particular details of our transfer learning methodology: we  review the fundamentals of the models that we are going to fine-tune (BERT and ClimateBert) and we describe the data used to fine-tune the models. In a following section –section 3–  we describe our proposed fine-tuning methodology of both the models - the BERT fine-tuning methodology that we use as a baseline and our proposed fine-tuning methodology of the ClimateBERT model. Then, in section, 4 we will analyze the obtained results in the illustrative experiments. Finally, a conclusions and further work section closes our paper. 

\section{Fundamentals of large language models for supervised classification}

Natural language processing (NLP) is a multidisciplinary area coming from the intersection of artificial intelligence and linguistics that studies methods to process and analyze large amounts of natural language data \cite{nadkarni2011natural}. In this context, we can fine-tune large language models, like transformers such as BERT (Bidirectional Encoder Representations from Transformers) or the GPT (Generative pretrained transformers) family, to efficiently solve a task that is similar to the one that they were trained on, which is precisely the methodology that we follow in this paper with the ClimateBert model. In this section, we first show the fundamentals of transfer learning to understand the motivation behind choosing this methodology and then we describe the particular details of our transfer learning methodology.  

In particular, one of the tasks addresed by NLP is to classify the texts according to a particular label based on a labeled corpus of texts. For example, to determine whether the sentiment of a particular social network publication is positive or negative based on previous labeled social network publications. Concretely, this action is known as supervised learning classification \cite{murphy2012machine} and has been typically solved by machine learning methods that used the different words of the texts as features. However, the field is experiencing an enormous revolution since the implementation of transformer models has been possible due to the rise of computing. Specifically, empirical results shown by transformers dramatically outperform the classical pipeline of machine learning models with a bag-of-words representation of the most common and relevant words of the texts according to algorithms such as TF-IDF \cite{gonzalez2020comparing}. Driven by the amazing results of transformer models, several communities like finance \cite{zhao2021bert}, energy \cite{cai2020sentiment} or tourism \cite{chantrapornchai2021information} are starting to use them for a wide variety of tasks.  

Transformer models are high capacity deep learning models, meaning that they have a plethora of parameters to be optimized, that need an enormous text corpora to estimate the value of their parameters in order to minimize the estimation of the generalization error of its predictions \cite{lecun2015deep}. Consequently, they can only be accurately estimated by supercomputing centers or organizations that have lots of computing power, being impossible to be trained from zero by small organizations or businesses for small supervised learning tasks. However, although the capacity issue seems to be a major limitation for its application in research problems such as the one targeted by this manuscript, it is eventually not. Most interestingly, we can use transfer learning to fine-tune a transformer trained in a similar task to the one that needs to be solved. In order to understand why the fine-tuning of the last layers of a deep neural network works, it is important to remark that the corpora where the parameters of the transformers are estimated usually contain huge sources of information such as Twitter, Github or online newspapers. Hence, depending on the problem, it is possible that a transformer creates a model of language from all the relevant sources of information, that we can consider as a relevant population for a study. Consequently, it is possible to download open source transformers that solve tasks as sentiment analysis and fine-tune them to a particular problem. Some model sources to download transformers include HuggingFace or Github, that are public repositories where we can download these transformers to use them in practice. 

The fine-tuning process adapts the behaviour of the transformer to the particular task to be solved and, moreover, it is cheap in computational terms, thus, allowing any organization to perform it. Besides, the results that a fine-tuned transformer can deliver outperform classical methodologies, or small models trained from zero \cite{gonzalez2020comparing} when the task to be solved is very similar to the task that the original transformer solves. When fine-tuning is performed to a big model to solve a similar task we call this process transfer learning. As we need to solve a supervised learning problem using a fine-tuned big model we can perform transfer learning to efficiently solve our task. 

Nowadays, there is a plethora of large language models, mainly transformers, that differ on their architecture and task solved. Concretely, transformers were first implemented on 2017, as a result of the prestigious paper Attention is all you need \cite{vaswani2017attention}, where the architecture of BERT was illustrated. In the next section, 2.1., we describe the particular details of the transformer that we use, that is related with the BERT transformer. 

\subsection{The ClimateBert model}

As we already pointed out, in recent years, a growing literature has emerged that relies on AI for the identification of climate-related information. More concretely, BoW (Bag of Words)-based approaches have been predominant, although this technique has the significant drawback of assuming that the words in the text are independent of one another, which is evidently false. On the other hand, transformers are large language models that capture the dependencies between words, that are encoded in word embeddings whose space represents the meaning of the words. 

Consequently, we use a transformer model, a BERT-related model called ClimateBert, that has been specifically tailored for climate text classification.  This model reads a whole sequence of words at once, allowing it to learn the context of a word from its surroundings. More specifically, ClimateBert is the state-of-the-art NLP model based on the Transformer architecture which has been specifically pre-trained on climate-related text corpora of over 1.6 billion paragraphs in relation to climate change consisting on news articles, corporate climate reports and research abstracts \cite{webersinke2021climatebert}. 

Regarding our research, one of the most significant breakthroughs of ClimateBert is the domain adaptive pre-training, being the first climate domain adaptive pre-trained model which has been made available to the public. In this sense, the model was pre-trained in the following NLP downstream tasks: text classification, sentiment analysis of risk and opportunity of corporations’ statements and fact-checking climate-related claims. In this sense, in the present work we have undertaken the ClimateBert fine-tuning process on the ClimaText corpus (described in next section 2.2), related to climate change disclosure, in order to compare the results obtained in the NLP text classification task using fine-tuning of the ClimateBert model with those provided by the BERT model, serving as a baseline because it achieves better performance that the standard BoW-machine learning architecture \cite{gonzalez2020comparing}. 

\subsection{Description of the data used to fine-tune the model}

For the purpose of conducting the present study and fine-tuning the ClimateBert model, we use the database named ClimaText \cite{varini2020climatext}. ClimaText is a dataset that intends to advance in the identification of climate change related topics in text corpora through NLP techniques. The data that compose this dataset represent sentences which are labeled depending on whether they talk about climate change (in which case, they will be labeled with a 1) or not (labeled with a 0). In particular, several sources were employed to create the database, such as Wikipedia, 10-K filings from the U.S. Securities and Exchange Commission (hereinafter, SEC), and a selection of climate change-related claims gathered from the web. 

In this regard, in order to train both models (the BERT fine-tuned and the ClimateBert fine-tuned), we have used the "AL-Wiki" dataset formed by a group of sentences extracted from Wikipedia to which we have applied DUALIST, an interactive machine learning (hereinafter, AL) system that enables the building of classifiers for data processing tasks. Concretely, the AL training sentences dataset thus consists of 3,000 sentences, of which 261 are related to the topic of climate change as opposed to 2,739 that are not, making it a tremendously unbalanced dataset. Furthermore, the test set used in the model is the "10 Ks (2018, test)", consisting of sentences that are extracted from Item 1A of the 2018 10-K Filings. It includes 300 sentences of which 67 address climate change compared to 233 that do not.  

It is important to emphasize the relevance of assessing the accuracy of both text classification models using data from the 10-K filings. These reports are required by the SEC for publicly listed US companies, which are obliged to provide a detailed annual presentation of their financial results, more specifically the company's history and activity, its financial statements, the risks it faces, its earnings per share, and any other relevant data. For all these reasons, 10-K Filings provide a very useful decision-making tool for investors. Since 2006, the SEC has required 10-K Filings to include Item 1A. For this purpose, Item 1A, on the basis of which the evaluation set of both models have been created, is known as "Risk Factors", thus including the most relevant risks faced by the company, which may be valid for the entire economy, for the industrial sector in which the firm operates, for a specific geographic area or exclusively for that company (SEC, 2021).    

In this manner, the SEC has adopted a so-called principles-based approach, whereby companies must self-identify the climate change-related risks to their business and disclose these risks in Item 1A of the 10-K filings. It is a significant advantage within the SEC's approach that disclosure of such information is mandatory and that failure to disclose climate risks can result in litigation with U.S. authorities \cite{kolbel2022ask}   

Additional advantages to using this data include the fact that the regulatory reports reflect both physical and transitional risks that are associated with climate change. It should be noted that the 10-K filings provide a description of future risks to the company and may therefore represent a better indicator of the future exposure that the company will face. This forward-looking nature is deliberate as investors often seek to obtain and incorporate forward-looking financial information \cite{kolbel2022ask}. 

Lastly, prior to the explanation of our methodology, it is important to clarify what is meant by climate change in the database, that is to say, when a sentence is being classified with a 1. In this sense, the labelling rules to classify a sentence as positive or 1 followed are as follows: the sentence must talk about climate change. To be more precisely, simply discussing the topic of environment is not enough. Moreover, if the sentence is describing a general scientific or climate fact it will only be relevant in the event that is a cause or effect of climate change (for example, while the sentence “Methane is CH4” will not be labeled as positive, “Methane increases temperature” will be). Those sentences that speak about clean energy, fossil fuels or emissions, among others, need to be connected to a societal or environmental aspect of climate change. Additionally, environmental aspects such as acid rain or pollution are not considered in this dataset as climate change related terms and the use of acronyms or names of entities (such as EPA -Environmental Protection Agency-) need to be mentioned with a cause or effect of climate change. Also, the sentence can discuss climate change during any time period. If there is any doubt and in all other cases, the sentence is labeled as negative or 0.   

\section{Proposed Methodology: Fine tuning of BERT and ClimateBERT}

Having reviewed the fundamentals of the models that we are going to fine-tune (BERT and ClimateBert) and having illustrated the data, we now describe the fine-tuning process of the models. Recall that we target a binary classification problem where we want to predict whether a text has climate change as a topic or not, according to the previous section. We first give the details of the BERT fine-tuning methodology that we use as a baseline and then those of our proposed fine-tuning methodology of the ClimateBERT model. Then, in a further section, we will analyze the obtained results. 

The pre-trained BERT model has been loaded from the TensorFlow Hub repository in order to fine-tune it to our database. The text entries have been set to lowercase prior to carrying out their tokenization, as well as the accents have been removed. We are working with a training batch size of 32 and setting the maximum length of the input token sequences to 128, typical values set by the community. Concretely, the model has been trained by means of fine-tuning using 3 epochs on the Wikipedia training dataset, previously described. 

The pre-trained ClimateBert model has been loaded from HuggingFace repository \cite{wolf2019huggingface}, in order to fine tune it to our database. In this sense, ClimateBert’s tokenizer does not need for the text entries to be previously set into lowercase or for the accents to be removed. We use the default hyper-parameter values of the Trainer library, being aware that an intelligent hyper-parameter tuning of these values, for instance using Bayesian optimization \cite{garrido2021advanced}, would enhance the predictive out-of-sample performance of the classifier. Lastly, for a fair comparison with the baseline, we also use 3 epochs for fine-tuning.  

\section{Illustrative experiments and results}

In order to estimate the error of generalization of the models regarding their climate change prediction, we use the accuracy metric estimator as we do not give more weight to false positives or false negatives. Concretely, the accuracy of the model indicates the total number of predictions that are correctly predicted with respect to the overall number of predictions performed by the model. As explained in the prior section, our training data is highly unbalanced, so it is necessary that the accuracy achieved by the validation surpasses, at least, the majority rule. Thus, as long as the sentences labeled 0 represent the majority of the training data set - namely 91.3\% - the accuracy achieved must be ensured to be above this value. 

For a single experiment, the BERT baseline  accomplished an accuracy of 95.7\% of the validation set which has been reached with the third epoch of fine-tuning. As in the training set the accuracy is slighlty higher, we stop the fine-tune process to avoid overfitting. On the other hand, we obtain an accuracy point estimation of 97.04\% using our refined version of ClimateBert, outperforming BERT’s performance with the same epochs of fine-tuning and the same setting. 

Subsequently, we proceed to evaluate the BERT model using the 2018 10-K Filings test dataset. The model trained by fine-tuning BERT has been able to achieve a significant performance, obtaining a point estimation of the accuracy of 90\%. Analogously, we proceed to evaluate the ClimateBert fine-tuned model using the same Filings test dataset. In this case, the model trained by fine-tuning ClimateBert has been able to outperform BERT, achieving a significant performance, reaching a  point estimation accuracy of 93\%. 

Having achieved these satisfactory statistical results, we proceeded once again to perform a hypothesis test to verify whether the results obtained by the model through the sample data can be considered as representative parameters, which is to say, we look to determine whether the results of the fine-tuned ClimateBert model for the classification of text on climate change are truly significant. 

For this purpose, we execute the ClimateBert and BERT models 25 times (to ensure the validity of the statistical hypothesis test as a consequence of the central limit theorem) with different random seeds and different splits of the train, validation and test sets. Using this methodology, we provide an interval estimation via a bootstrapping process of the unknown accuracy, precision, F1, recall and specificity parameters and also include information from the fine-tuned BERT carried out in the paper of Varini et al. (2020), obtaining the following results (see Table 1). 

\begin{table}
\begin{center}
\begin{tabular}{||c | c | c | c | c | c | c||} 
 \hline
 Model & Accuracy & Precision & F1 & Recall & Specificity & Deviation \\ [0.5ex] 
 \hline\hline
 Varini BERT fine-tuned & 0.83 & 0.58 & 0.71 & $\mathbf{0.94}$ & Unknown & Unknown \\ 
 \hline
 Our BERT fine-tuned & 0.925 & 0.93 & 0.91 & 0.93 & 0.747 & 0.006 \\
 \hline
 ClimateBERT fine-tuned & $\mathbf{0.935}$ & $\mathbf{0.94}$ & $\mathbf{0.93}$ & 0.94 & $\mathbf{0.786}$ & 0.007 \\
 [1ex] 
 \hline
\end{tabular}
\caption{Performance comparison between the methodologies.}
\end{center}
\end{table}

We can conclude that our fine-tuned model (ClimateBert) has better performance with respect to the baseline (fine-tuned BERT) due to the fact that if we compute a two sample t-test with pooled variance and a confidence level of 99\% with the previous data, we obtain a p-value of .000003154, representing that the null hypothesis of both models having the same performance is remote. 

Finally, it can be claimed that, with the sole exception of the metric of recall, the results we have obtained from the BERT fine tuning process to the aforementioned ClimaText sources have outperformed those previously obtained in \cite{varini2020climatext}. Moreover, these results have been further improved by using the state of the art algorithm, Climatebert, for the NLP task of climate-related text classification.  

All in all, it can be said that the fine-tuned ClimateBert model has got a greater capacity to predict not only true positives (sentences which are climate-related) but also to classify true negatives (sentences which are not speaking about climate change). In this sense, ClimateBert has outperformed BERT in the task of climate change text classification.   

\section{Conclusions and further research}

The present study has got the purpose to further contribute to the literature on the detection of the topic of climate change in text corpora, particularly in order to encourage the identification by economic agents of climate-related financial risks disclosure of firms. For this purpose, we have applied NLP techniques that enable us to carry out the text classification task with optimal results with high performance levels. In particular, our text classification model, obtained from the ClimateBert fine-tuning process, outperforms the models created with BERT, both our own and the one introduced by Varini et al \cite{varini2020climatext}.  We believe that this result truly represents a contribution to the literature regarding the detection of climate change in text corpora. 

Based on our results, it can be said that the ultimate team of ClimateBert and Climatext is an outstanding tool within the NLP pre-trained models that may and should be used by investors, institutional agents and companies themselves to monitor the disclosure of climate risk in financial reports, as well as in other textual sources, such as policies, new legislation or directives passed on this matter. Although in the present paper we have carried out a text classification task in order to compare the fine-tuned BERT and fine-tuned ClimateBert models, the door remains open for the joint use of  ClimaText and ClimateBert in other NLP tasks, such as sentiment analysis. Likewise, another area of further research is to jointly apply ClimateBert and ClimaText in order to analyze wheher the European companies, have significantly improved their climate-related disclosures after the mandatory request of the European Non-Financial Reporting Directive (NFRD). 

\bibliography{main}
\bibliographystyle{acm}

\end{document}